\pdfoutput=1

\documentclass[11pt]{article}

\usepackage[]{acl}

\usepackage{times}
\usepackage{latexsym}

\usepackage[T1]{fontenc}

\usepackage[utf8]{inputenc}

\usepackage{microtype}

\usepackage{microtype}
\usepackage{graphicx}
\usepackage{booktabs}
\usepackage{subcaption}
\usepackage{amsmath}
\usepackage{multirow}
\usepackage{enumitem}

\usepackage[T2A,T1]{fontenc}
\usepackage[utf8]{inputenc}
\usepackage[russian,english]{babel}
\usepackage{newtxmath}

\newcommand{\method}{\textsc{DEEP}\xspace}
\newcommand{\secref}[1]{\S\ref{sec:#1}}
\newcommand\boldblue[1]{\textcolor{blue}{\textbf{#1}}}
\newcommand\blue[1]{\textcolor{blue}{#1}}
\newcommand\red[1]{\textcolor{red}{#1}}
\newcommand{\boldred}[1]{\textcolor{red}{\textbf{#1}}}
\newcommand{\loss}[1]{\mathcal{L}_{\text{#1}}}
\newcommand{\circled}[1]{\textcircled{\raisebox{-0.9pt}{#1}}}
\newcommand{\given}{\,|\,}
\newcommand*{\affmark}[1][*]{\textsuperscript{#1}}

\usepackage{xspace,mfirstuc,tabulary}

\title{DEEP: DEnoising Entity Pre-training for Neural Machine Translation}

\author{
 Junjie Hu\affmark[1], Hiroaki Hayashi\affmark[3], Kyunghyun Cho\affmark[2], Graham Neubig\affmark[3]\\
\affmark[1]University of Wisconsin-Madison, \affmark[2]New York University, \affmark[3]Carnegie Mellon University \\
  {\sf junjie.hu@wisc.edu, kyunghyun.cho@nyu.edu, \{hiroakih,gneubig\}@cs.cmu.edu} \\
}

\date{}

\begin{document}
\maketitle

\begin{abstract}
It has been shown that machine translation models usually generate poor translations for named entities that are infrequent in the training corpus. Earlier named entity translation methods mainly focus on phonetic transliteration, which ignores the sentence context for translation and is limited in domain and language coverage. To address this limitation, we propose \method, a \textbf{DE}noising \textbf{E}ntity \textbf{P}re-training method that leverages large amounts of monolingual data and a knowledge base to improve named entity translation accuracy within sentences. Besides, we investigate a multi-task learning strategy that finetunes a pre-trained neural machine translation model on both entity-augmented monolingual data and parallel data to further improve entity translation. Experimental results on three language pairs demonstrate that \method results in significant improvements over strong denoising auto-encoding baselines, with a gain of up to 1.3 BLEU and up to 9.2 entity accuracy points for English-Russian translation.

\end{abstract}

\section{Introduction}

\begin{figure*}[tb]
    \centering
    \includegraphics[width=\textwidth]{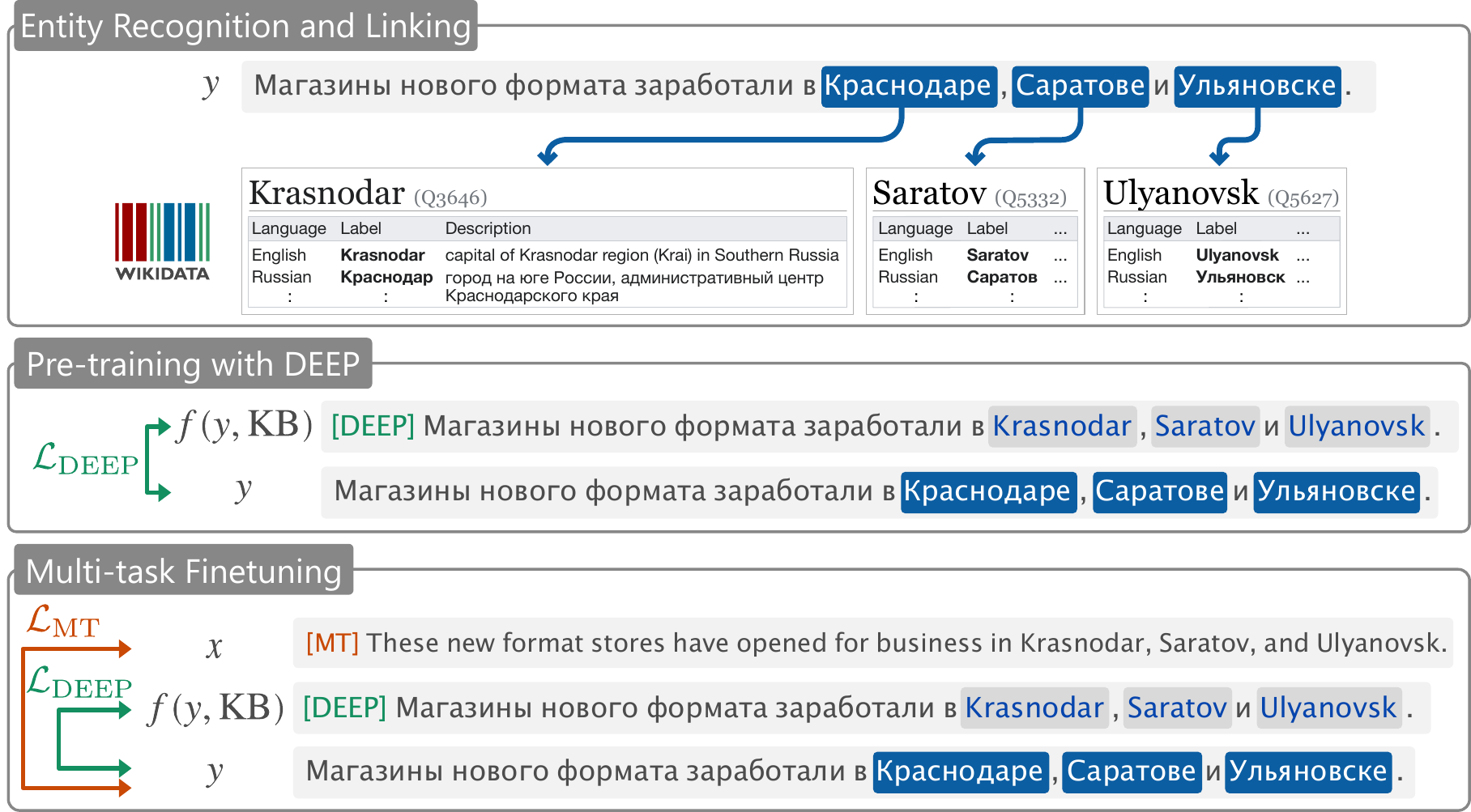}
    \caption{General workflow of our method. Entities in a sentence is extracted and linked to Wikidata, which includes their translations in many languages. \method uses the noise function $f(y, \text{KB})$ that replaces entities with the translations for pre-training. \method is also employed during finetuning in a multi-task learning manner.}
    \label{fig:workflow}
\end{figure*}

Proper translation of named entities is critically important for accurately conveying the content of text in a number of domains, such as news or encyclopedic text~\cite{knight-graehl-1998-machine,al2002named,al-onaizan-knight-2002-translating}. 
In addition, a growing number of new named entities (e.g., person name, location) appear every day, and as a consequence many of these entities may not exist in the parallel data traditionally used to train MT systems.
As a consequence, even state-of-the-art MT systems struggle with entity translation.
For example, \citet{laubli2020jair} note that a Chinese-English news translation system that had allegedly reached human parity still lagged far behind human translators on accurate translation of entities, and this problem will be further exacerbated in the settings of cross-domain transfer or in the case of emerging entities.

Because of this, there have been a number of methods proposed specifically to address the problem of translating entities.
As noted by \citet{liu2015technical}, earlier studies on named entity translation largely focused on rule-based methods~\cite{wan-verspoor-1998-automatic}, statistical alignment methods~\cite{huang-etal-2003-automatic,huang-etal-2004-improving} and Web mining methods~\cite{huang-etal-2005-mining,wu-chang-2007-learning,yang-etal-2009-chinese}.
However, these methods have two main issues.
First, as they generally translate a single named entity without any context in a sentence, it makes it difficult to resolve ambiguity in entities using context.
In addition, the translation of entities is often performed in a two-step process of entity recognition then translation, which complicates the translation pipeline and can result in cascading errors~\cite{huang-etal-2003-automatic,huang-etal-2004-improving,chen-etal-2013-joint}. 

In this paper, we focus on a simple yet effective method that improves named entity translation within context.
Specifically, we do so by devising a data augmentation method that leverages two data sources: monolingual data from the target language and entity information from a knowledge base (KB).
Our method also adopts a procedure of pre-training and finetuning neural machine translation (NMT) models that is used by many recent works~\cite{Luong-Manning:iwslt15,neubig-hu-2018-rapid,pmlr-v97-song19d,liu-etal-2020-multilingual-denoising}. In particular, pre-training methods that use monolingual data to improve translation for low-resource and medium-resource languages mainly rely on a denoising auto-encoding objective that attempt to reconstruct parts of text~\cite{pmlr-v97-song19d} or the whole sentences~\cite{liu-etal-2020-multilingual-denoising} from noised input sentences without particularly distinguishing named entities and other functional words in the sentences. In contrast, our method exploits an entity linker to identify entity spans in the monolingual sentences and link them to a KB (such as Wikidata~\cite{vrandevcic2014wikidata}) that contains multilingual translations of these entities. We then generate noised sentences by replacing the extracted entity spans with their translations in the knowledge base and pre-train our NMT models to reconstruct the original sentences from the noised sentences. To further improve the entity translation accuracy and avoid forgetting the knowledge learned from pre-training, we also examine a multi-task learning strategy that finetunes the NMT model using both the denoising task on the monolingual data and the translation task on the parallel data.

In the experiments on English-Russian, English-Ukrainian and English-Nepali translations, \method outperforms the strong denoising auto-encoding baseline with respect to entity translation accuracy, and obtains comparable or slightly better overall translation accuracy as measured by BLEU. A fine-grained analysis shows that our multi-task finetuning strategy improves the translation accuracy of the entities that do not exist in the finetuning data.

\section{Denoising Auto-Encoding (DAE)}
Given a set of monolingual text segments for pre-training, i.e., $y\in\mathcal{D}_Y$, a sequence-to-sequence denoising auto-encoder is pre-trained to reconstruct a text segment $y$ from its noised version corrupted by a noise function $g(\cdot)$. Formally, the DAE objective is defined as follows:
\begin{align}
    \loss{DAE}(\mathcal{D}_Y, \theta) = \sum_{y\in\mathcal{D}_Y} \log P(y \given g(y); \theta),
\end{align}
where $\theta$ denotes the model's learning parameters. For notation simplicity, we drop $\theta$ in the rest of the sections.
This formulation encompasses several different previous works in data augmentation for MT, such as monolingual data copying \citep{currey-etal-2017-copied}, where $g(\cdot)$ is the identity function, back translation \citep{sennrich-etal-2016-improving}, where $g(\cdot)$ is a backwards translation model, as well as heuristic noising functions \citep{pmlr-v97-song19d,lewis-etal-2020-bart,liu-etal-2020-multilingual-denoising} that randomly sample noise according to manually devised heuristics.

In particular, as our baseline we focus on the mBART method \citep{liu-etal-2020-multilingual-denoising}, a recently popular method with two type of heuristic noise functions being used sequentially on each text segment. The first noise function randomly masks spans of text in each sentence. Specifically, a span length is first randomly sampled from a Poisson distribution ($\lambda=0.35$) and the beginning location for a span in $y$ is also randomly sampled. The selected span of text is replaced by a mask token. This process repeats until 35\% of words in the sentence are masked. The second noise function is to permute the sentence order in each text segment with a probability.

\section{Denoising Entity Pre-training} \label{sec:deep}

Our method adopts a procedure of pre-training and finetuning for neural machine translation. First, we apply an entity linker to identify entities in a monolingual corpus and link them to a knowledge base (\secref{entity_linking}). We then utilize entity translations in the knowledge base to create noisy code-switched data for pre-training (\secref{pretrain}). Finally, we examine a multi-task learning strategy to further improve the translation of low-frequency entities (\secref{finetune}).

\subsection{Entity Recognition and Linking} \label{sec:entity_linking}


The goal of this part is to identify entities in each monolingual segment and obtain their translations.
To this end, we use Wikidata~\cite{vrandevcic2014wikidata} a multilingual knowledge base that covers 94M entities.\footnote{Statistics as of June 14, 2021.}
Each entity is represented in surface forms from different languages in which a Wikipedia article exists.
Therefore, linking an entity mention $t$ in a target-language segment $y$ to an entity $e$ in Wikidata allows us to obtain the multilingual translations of the entity, that is,
\begin{align}
    \forall t \in y, \exists e \in \text{KB}: \, T_e = \text{surface}(e, \text{KB}), t \in T_e \nonumber
\end{align}
where $T_e$ denotes a set of multilingual surface forms of $e$.
We can define the translate operation as: $s = \text{lookup}(T_e, X)$ which simply looks for the surface form of $e$ in the source language $X$.
Note that this strategy relies on the fact that translations in higher-resource languages are included in $T_e$, which we adopt by using English in our experiments.
In general, however, $T_e$ does not universally cover all the languages of interest.
For entity recognition and linking, we use SLING~\cite{ringgaard2017sling},\footnote{https://github.com/google/sling.} which builds an entity linker for arbitrary languages available in Wikipedia.

\subsection{Entity-based Data Augmentation}
\label{sec:pretrain}

After obtaining entity translations from the KB, we attempt to explicitly incorporate these translations into the monolingual sentences for pre-training. To do so, we design a entity-based noise function that takes in a sentence $y$ and the KB, i.e., $f(y, \text{KB})$. First, we replace all detected entity spans in the sentence by their translations from the KB:
\begin{align}
    \text{replace}(y, \text{KB}) = \text{swap}(s, t, y),~ \forall t \in y 
\end{align}
where the swap() function swaps occurrences of one entity span $t$ in $y$ with its translation $s$ in the source language.
For example, in the second box of Figure~\ref{fig:workflow}, the named entities ``\foreignlanguage{russian}{\blue{Краснодаре}, \blue{Саратове} and \blue{Ульяновске}}'' in Russian are replaced by their English translations ``\blue{Krasnodar}, \blue{Saratov}, and \blue{Ulyanovsk}''. After the replacement, we create a noised code-switched segment which explicitly include the translations of named entities in the context of the target language. For some segments that contain fewer entities, their code-switched segments may be similar to them, which potentially results in a easier denoising task. Therefore, we further add noise to these code-switched segments. To do so, if the word count of the replaced entity spans is less than a fraction (35\%) of the word count in the segment, we then randomly mask the other non-entity words to make sure that about 35\% of the words are either replaced or masked in the noised segment. 
Finally, we follow \citet{liu-etal-2020-multilingual-denoising} to randomly permute the sentence order in $y$. We then train a sequence-to-sequence model to reconstruct the original sentence $y$ from its noised code-switched sentence as follows: 
\begin{align}
    \loss{DEEP}(\mathcal{D}_Y, \text{KB})= \sum_{y\in\mathcal{D}_Y} \log P(y \given f(y,\text{KB})) \nonumber
\end{align}




\subsection{Multi-task Finetuning}
\label{sec:finetune}

After pre-training, we continue finetuning the pre-trained model on a parallel corpus $(x,y) \in \mathcal{D}_{XY}$ for machine translation.

\begin{align}
    \loss{MT}(\mathcal{D}_{XY}) = \sum_{(x,y) \in \mathcal{D}_{XY}} \log P(y \given x)
\end{align}

To avoid forgetting the entity information learned from the pre-training stage, we examine a mutlitask learning strategy to train the model by both the pre-training objective on the monolingual data and the translation objective on the parallel data. Since monolingual segments are longer text sequences than sentences in $\mathcal{D}_{XY}$ and the size of $\mathcal{D}_{Y}$ is usually larger than that of $\mathcal{D}_{XY}$, simply concatenating both data for multi-task finetuning leads to bias toward denoising longer sequences rather than actually translating sentences. To balance the two tasks, in each epoch we randomly sample a subset of monolingual segments $\mathcal{D}_{Y}'$ from $\mathcal{D}_{Y}$, where the total subword count of $\mathcal{D}_{Y}'$ equals to that of $\mathcal{D}_{XY}$, i.e., $\sum_{y\in\mathcal{D}_y'}|y| = \sum_{(x,y)\in\mathcal{D}_{XY}}\max(|x|,|y|) $. We then examine the multitask finetuning as follows:
\begin{align}
    \loss{Multi-task} = \loss{MT}(\mathcal{D}_{XY}) +  \loss{Pre-train}(\mathcal{D}_Y')
\end{align}
where the pre-training objective $\loss{Pre-train}$ is either DAE or \method with \method having an additional input of a knowledge base. 
Notice that with the sampling strategy for the monolingual data, we double the batch size in the multi-task finetuning setting with respect to that in the single-task finetuning setting. Therefore, we make sure that the models are finetuned on the same amount of parallel data in both the single-task and multi-task settings, and the gains from the mutlitask setting sorely come from the additional task on the monolingual data.

To distinguish the tasks during finetuning, we replace the start token (``[BOS]'') in a source sentence or a noised segment by the corresponding task tokens for the translation or denoising task (i.e., ``[MT]'', ``[DAE]'' or ``[DEEP]''). We initialize the additional task embeddings by the start token embedding and append these task embeddings to the word embedding matrix of the encoder.


\section{Experimental Setting}

\paragraph{Pre-training Data:} We conduct our experiments on three language pairs: English-Russian, English-Ukrainian and English-Nepali. We use Wikipedia articles as the monolingual data for pre-training and report the data statistics in Table~\ref{tab:pretrain_stats}. We tokenize the text using the same sentencepiece model as \citet{liu-etal-2020-multilingual-denoising}, and train on sequences of 512 subwords. 

\begin{table}[tb]
    \centering 
    \begin{tabular}{lrrrrr} \toprule
        \multirow{2}{*}{Lang.} &  \multirow{2}{*}{Token} &  \multirow{2}{*}{Para.} & \multicolumn{3}{c}{Entity} \\ \cmidrule{4-6}
        &  &  & Type & Count & N \\\midrule
         Ru & 775M & 1.8M & 1.4M & 337M & 123 \\
         Uk & 315M & 654K & 524K & 140M & 149\\
         Ne & 19M & 26K & 17K & 2M & 34 \\ \bottomrule
    \end{tabular} 
    \caption{Statistics of Wikipedia corpora in Russian (Ru), Ukrainian (Uk) and Nepali (Ne) for pre-training. $N$ denotes the average subword count of entity spans in a sequence of 512 subwords.}
    \label{tab:pretrain_stats}
    \vspace{-5mm}
\end{table}


\paragraph{Finetuning \& Test Data:} We use the news commentary data from the English-Russian translation task in WMT18 for finetuning and evaluate the performance on the WMT18 test data from the news domain. For English-Ukrainian, we use the TED Talk transcripts from July 2020 in the OPUS repository~\cite{tiedemann-2012-parallel} for finetuning and testing. For English-Nepali translation, we use the FLORES dataset in \citet{guzman-etal-2019-flores} and follow the paper's setting to finetune on parallel data in the OPUS repository. Table~\ref{tab:finetune_data} shows the data statistics of the parallel data for finetuning. Notice that from the last four columns of Table~\ref{tab:finetune_data}, the entities in the pre-training data cover at least 87\% of the entity types and 91\% of the entity counts in both finetuning and test data except the En-Ne pair.

\begin{table}[tb]
    \centering \setlength{\tabcolsep}{3pt}
    \resizebox{0.49\textwidth}{!}{%
    \begin{tabular}{lccccccc} \toprule
        \multirow{2}{*}{Lang.} &  \multirow{2}{*}{Train} &  \multirow{2}{*}{Dev} & \multirow{2}{*}{Test} & \multicolumn{2}{c}{PF / F} &\multicolumn{2}{c}{PT / T}  \\ \cmidrule{5-8}
        &  &  & & Type & Count & Type & Count \\\midrule
         En-Ru & 235K & 3.0K & 3.0K & 88\% & 94\% & 88\% & 91\% \\
         En-Uk & 200K & 2.3K & 2.5K & 87\% & 94\% & 91\% & 94\% \\
         En-Ne & 563K & 2.6K & 2.8K & 35\% & 25\% & 44\% & 27\%\\ \bottomrule
    \end{tabular} }
    \caption{Statistics of the parallel train/dev/test data for finetuning. Type and Count under PF/F (PT/T) show the percentage of entity types and counts in the finetuning (test) data that are covered by the pre-training data.  }
    \label{tab:finetune_data}
    \vspace{-5mm}
\end{table}

\paragraph{Architecture:} We use a standard sequence-to-sequence Transformer model~\cite{NIPS2017_3f5ee243} with 12 layers each for the encoder and decoder. We use a hidden unit size of 512 and 12 attention heads. Following~\citet{liu-etal-2020-multilingual-denoising}, we add an additional layer-normalization layer on top of both the encoder and decoder to stabilize training at FP16 precision. We use the same sentencepiece model and the vocabulary from~\citet{liu-etal-2020-multilingual-denoising}.

\paragraph{Methods in Comparison:} We compare methods in the single task and multi-task setting as follows:
\begin{itemize}[leftmargin=10pt]\itemsep-0.2em
    \item \textbf{Random $\rightarrow$ MT}: We include a comparison with a randomly initialized model without pre-training and finetune the model for each translation task.
    \item \textbf{DAE $\rightarrow$ MT}: We pre-train a model by DAE using the two noising functions in~\citet{liu-etal-2020-multilingual-denoising} and finetune the model for each translation task.
    \item \textbf{DEEP $\rightarrow$ MT}: We pre-train a model using our proposed DEEP objective and finetune the model on the translation task.
    \item \textbf{DAE $\rightarrow$ DAE+MT}: We pre-train a model by the DAE objective and finetune the model for both the DAE task and translation task.
    \item \textbf{DEEP $\rightarrow$ DEEP+MT}: We pre-train a model by the DEEP objective and finetune the model for both the DEEP task and translation task.
\end{itemize}

\paragraph{Learning \& Decoding:} We pre-train all models for 50K steps first using the default parameters in \citet{liu-etal-2020-multilingual-denoising} except that we use a smaller batch of 64 text segments, each of which has 512 subwords. We use the Adam optimizer ($\epsilon$=1e-6, $\beta_2$=0.98) and a polynomial learning rate decay scheduling with a maximum step at 500K. All models are pre-trained on one TPUv3 (128GB) for approximately 12 hours for 50K steps.\footnote{As we show in Figure~\ref{fig:pretrain_vs_perf}, models pre-trained for 50K steps have provided a reasonably good initialization.} 
We then reset the learning rate scheduler and continue finetuning our pre-trained models on the MT parallel data for 40K steps. 
We set the maximum number of tokens in each batch to 65,536 in the single task setting and double the batch size in the multi-task setting. 
We use 2,500 warm-up steps to reach a maximum learning rate of 3e-5, and use 0.3 dropout and 0.2 label smoothing. After training, we use beam search with a beam size of 5 and report the results in BLEU following the evaluation in~\citet{liu-etal-2020-multilingual-denoising}.

\section{Discussion}

\subsection{Corpus-level Evaluation}
In Table~\ref{tab:BLEU}, we compare all methods in terms of BLEU on the test data for three language pairs. First, we find that all pre-training methods significantly outperform the random baseline. In particular, our \method method obtains a substantial gain of 3.5 BLEU points in the single task setting for the low-resource En-Ne translation. Second, we observe improvements with the multi-task finetuning strategy over the single-task finetuning for all language pairs. Our \method method outperforms the DAE method for En-Ru translation by 1.3 BLEU points in the multi-task setting. It is also worth noting that \method obtains higher BLEU points than DAE at the beginning of the multi-task finetuning process, however the gap between both methods decreases as the finetuning proceeds for longer steps (See Appendix~\ref{sec:stepwise_finetune}). One possible reason is that models trained by \method benefit from the entity translations in the pre-training data and obtain a good initialization for translation at the beginning of the finetuning stage. As the multitask finetuning proceeds, the models trained by both DAE and \method rely more on the translation task than the denoising task for translating a whole sentence. Thus the nuance of the entity translations might not be clearly evaluated according to BLEU.

\subsection{Entity Translation Accuracy}

Since corpus-level metrics like BLEU might not necessarily reveal the subtlety of named entity translations, in the section we perform a fine-grained evaluation by the entity translation accuracy which counts the proportion of entities correctly translated in the hypotheses. Specifically, we first use SLING to extract entities for each pair of a reference and a hypothesis. We then count the translation accuracy of an entity as the proportion of correctly mentioning the right entity in the hypotheses, followed by macro-averaging to obtain the average entity translation accuracy.
We show the results in Table~\ref{tab:translation_accuracy}.
First, our method in both single- and multi-task settings significantly outperformed the other baselines. In particular, the gains from \method are much clear for the En-Uk and En-Ru translations. One possible reason is that Russian or Ukrainian entities extracted from the pre-training data have a relatively higher coverage of the entities in both the finetuning and test data as reported in Table~\ref{tab:finetune_data}. However, SLING might not detect as many entities in Nepali as in the other languages. We believe that future advances on entity linking in low-resource languages could potentially improve the performance of \method further. We leave this as our future work.


\begin{table}[tb]
\centering
\resizebox{0.48\textwidth}{!}{%
    \begin{tabular}{lccc}
    \toprule
    Pre-train $\rightarrow$ Finetune & En-Uk & En-Ru & En-Ne \\ \midrule
    Random $\rightarrow$ MT & 17.1 & 15.0 & \phantom{0}7.7 \\
    DAE $\rightarrow$ MT & 19.5 & 18.5 & 10.5 \\
    DEEP $\rightarrow$ MT & 19.4 & 18.5  & 11.2 \\ \midrule
    DAE $\rightarrow$ DAE+MT & \textbf{19.7}  & 18.9 & \textbf{11.6} \\
    DEEP $\rightarrow$ DEEP+MT & \textbf{19.7}  & \textbf{19.6} &  11.5 \\ \bottomrule
    \end{tabular}%
}
\caption{BLEU in single- and multi-task settings.}
\label{tab:BLEU}
\end{table}

\begin{table}[tb]
\centering
\resizebox{0.48\textwidth}{!}{%
    \begin{tabular}{lccc}
    \toprule
    Pre-train $\rightarrow$ Finetune & En-Uk & En-Ru & En-Ne \\ \midrule
    Random $\rightarrow$ MT & 49.5	& 31.1	& 20.9 \\
    DAE $\rightarrow$ MT & 56.7	& 37.7	& 26.0 \\
    DEEP $\rightarrow$ MT & 57.7 & 	40.6 &	\textbf{28.6}\\ \midrule
    DAE $\rightarrow$ DAE+MT & 58.8 & 47.2 & 	27.9 \\
    DEEP $\rightarrow$ DEEP+MT & \textbf{61.9}	& \textbf{56.4} & 28.3 \\ \bottomrule
    \end{tabular}%
}
\caption{Entity translation accuracy in single- and multi-task settings.}
\label{tab:translation_accuracy}
\end{table}

\begin{figure*}
    \centering
    \includegraphics[width=\textwidth]{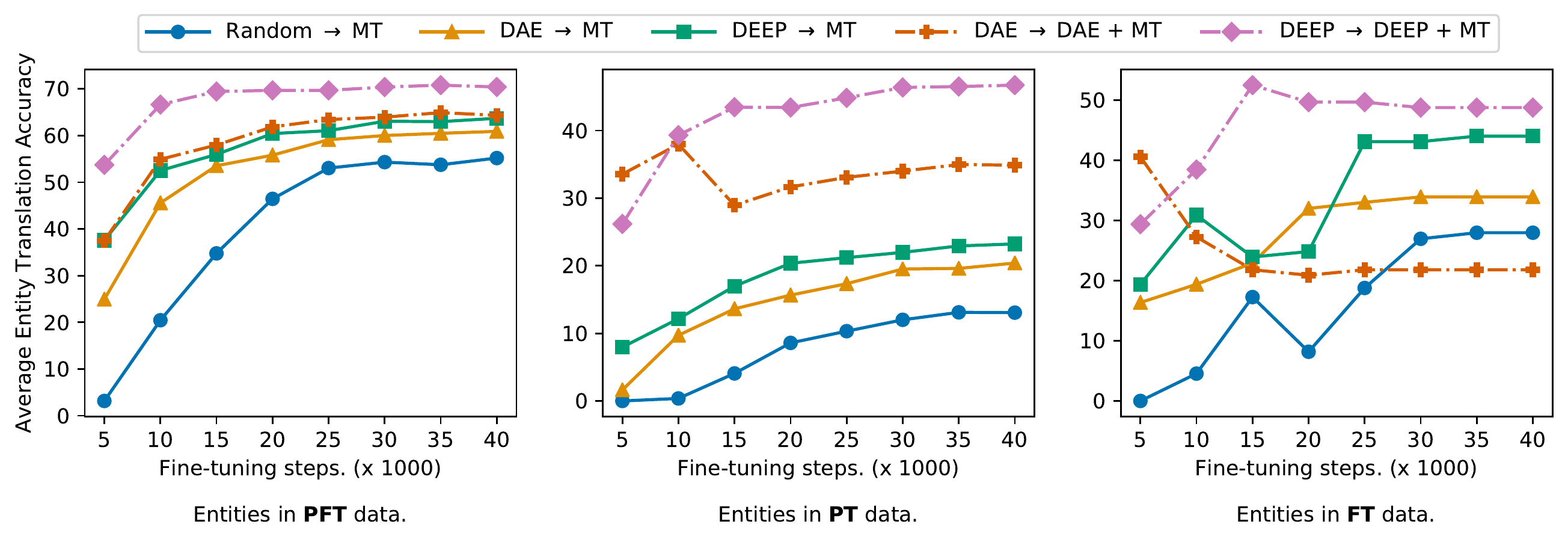}
    \caption{Entity translation accuracy scores aggregated over different entity sets for Russian. \textbf{PFT}, \textbf{PT}, \textbf{FT} data correspond to entities appearing in (i) pre-training, finetuning and test data, (ii) only pre-training and test data (iii) only finetuning and test data.}
    \label{fig:per_step_accuracy_ru}
    \vspace{-4mm}
\end{figure*}

\subsection{Fine-grained Analysis on Entity Translation Accuracy}


In this section, we further analyze the effect on different categories of entities using our method.

\paragraph{Performance of Entity Groups over Finetuning:}
The model is exposed to some entities more often than others at different stages: pre-training, finetuning and testing, which raises a question: \textit{how is the entity translation affected by the exposure during each stage?}
To answer this question, we divide the entities appearing in the test data into three groups:
\begin{itemize}[leftmargin=10pt]\itemsep-0.2em
    \item \textbf{PFT}: entities appearing in the pre-training, finetuning, and test data.
    \item \textbf{PT}: entities only in the pre-training and test data.
    \item \textbf{FT}: entities only in the finetuning and test data.
\end{itemize}
We show the English-to-Russian entity translation accuracy scores for each group over finetuning steps in Figure~\ref{fig:per_step_accuracy_ru}.
Overall, accuracies are higher for the entities that appear in the finetuning data (\textbf{PFT}, \textbf{FT}), which is due to the exposure to the finetuning data.
Our proposed method consistently outperformed baseline counterparts in both single- and multi-task settings.
The differences in accuracy are particularly large at earlier finetuning steps, which indicates the utility of our method in lower-resource settings with little finetuning data.
The effect of multi-task finetuning is most notable for entities in \textbf{PT}.
Multi-task finetuning continuously exposes the model to the pre-training data, which as a result prevents the model from forgetting the learned entity translations from \textbf{PT}.

\begin{figure*}
    \centering
    \includegraphics[width=\textwidth]{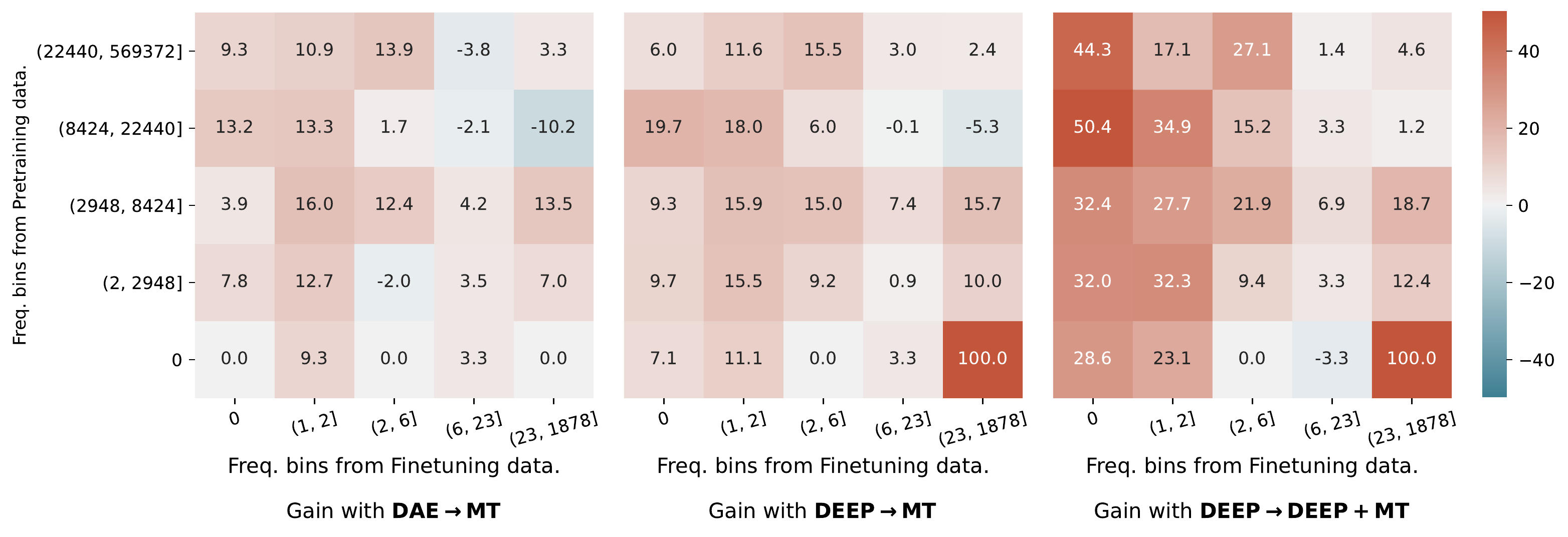}
    \caption{Gain from Random $\rightarrow$ MT in entity translation accuracy for each model.}
    \label{fig:ru_heatmap_40000_ent}
\end{figure*}

\paragraph{Performance according to Entity Frequency:}
We further analyze the entity translation accuracy scores using entity frequencies in each group introduced above.
This provides a more fine-grained perspective on \textit{how frequent or rare entities are translated}.
To do so, we take Russian hypotheses from a checkpoint with 40K steps of finetuning, bin the set of entities in three data (\textit{i.e.} \textbf{PFT}, \textbf{PT}, \textbf{FT}) according to frequencies in each of the data.
We then calculate the entity translation accuracy within each bin by comparing them against reference entities in the respective sentences.
Figure~\ref{fig:ru_heatmap_40000_ent} shows the accuracy gain of each pre-training methodologies from \textbf{Random $\rightarrow$ MT} (\textit{i.e.} no pre-training) on test data, grouped by the entity frequency bins in pre-training and finetuning data.
Note that leftmost column and the bottom row represent \textbf{PT}, \textbf{FT}, respectively.
As observed earlier, the proposed method improves more over most frequency bins, with greater differences on entities that are less frequent in finetuning data.
This tendency is observed more significantly for the multi-task variant (\textbf{DEEP $\rightarrow$ DEEP + MT}), where the gains are mostly from entities that never appeared in finetuning data (\textit{i.e.} leftmost column).
Multi-task learning with DEEP therefore prevents the model from forgetting the entity translations learned at pre-training time. Analytical results on Ukrainian and Nepali are in Appendix~\ref{app:eta_for_uk_ne}.

\subsection{Optimization Effects on DEEP}
\paragraph{Finetuning Data Size vs Entity Translation:} While \method primarily focuses on the application in a low-resource setting, the evaluation with more resources can highlight potential use in broader scenarios.
To this end, we expand the finetuning data for English-Russian translation with an additional 4 million sentence pairs from ParaCrawl \cite{banon-etal-2020-paracrawl}, a parallel data collected from web pages.
Although web pages might contain news text, the ParaCrawl data cover more general domains.
We finetune models on the combined data and evaluate with BLEU and entity translation accuracy.
Table~\ref{tab:finetune_size} shows the model comparisons across different finetuning data sizes.
When the model is initialized with pre-training methods, we observed decreased BLEU points and the increased entity translation accuracy scores.
On the one hand, this is partly due to the discrepancy in terms of domains between our finetuning data (news) and ParaCrawl.
Regardless, DEEP is consistently equal to or better than DAE in all tested settings.

\begin{table}[tb]
    \small
    \centering
    \begin{tabular}{lccccc} \toprule
        \multirow{2}{*}{Methods} & \multicolumn{2}{c}{0.24M} & & \multicolumn{2}{c}{4.25M}  \\ \cmidrule{2-3}\cmidrule{5-6}
         & BLEU & Acc. & & BLEU & Acc. \\ \midrule
         Random $\rightarrow$ MT & 15.0 & 31.1  &  & 15.7 & 39.4 \\
         DAE $\rightarrow$ MT    & 18.5 & 37.7  &  & 16.3 & 53.7 \\
         DEEP $\rightarrow$ MT   & 18.5 & 40.6  &  & 17.2 & 53.9 \\ \bottomrule
    \end{tabular}
    \caption{Model comparisons across different finetuning data sizes. The results on the right are obtained after finetuning on the combined news commentary and ParaCrawl data.}
    \label{tab:finetune_size}
\end{table}

\begin{figure}[tb]
    \centering
    \includegraphics[width=0.48\textwidth]{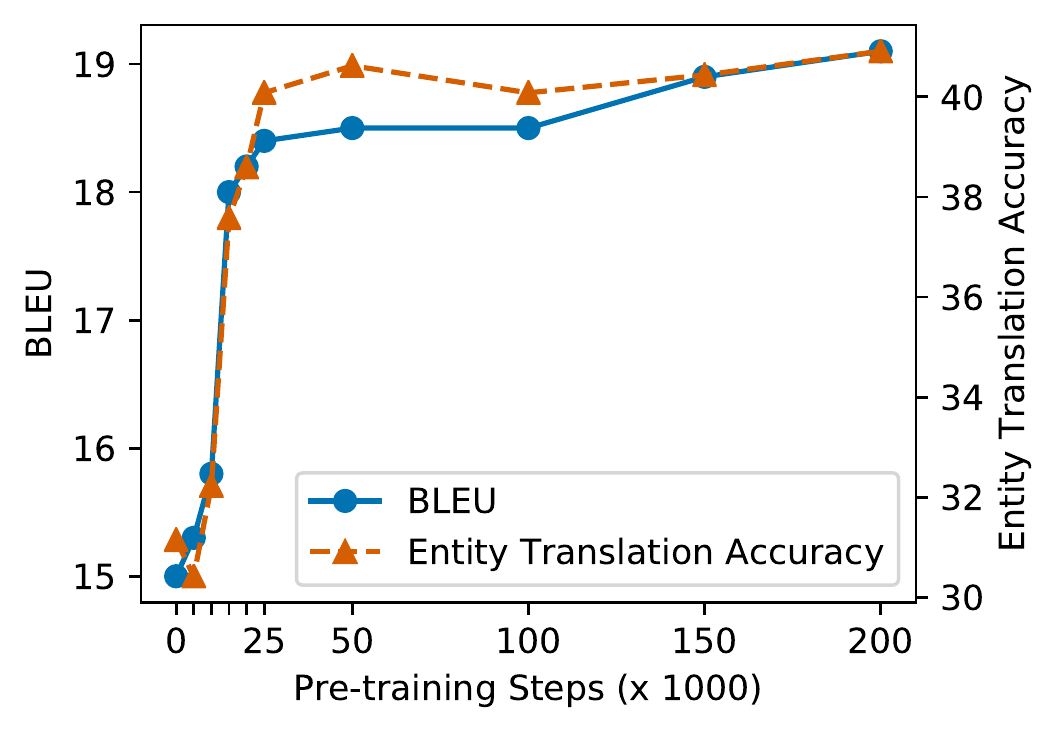}
    \caption{English-to-Russian BLEU and Entity translation accuracy scores after finetuning with respect to variable pre-training steps. Finetuning is performed for 40K steps.}
    \label{fig:pretrain_vs_perf}
\end{figure}

\paragraph{Pre-training Steps vs Entity Translation:}
Since \method leverages entity-augmented monolingual data, the model trained by \method revisits more entities in different context as the pre-training steps increase. To analyze the efficiency of learning name entity translations during pre-training stage, we focus on the question: \textit{how many pre-training steps are needed for named entity translation?} To examine this question, we take the saved checkpoints trained by \method from various pre-training steps, and apply the single-task finetuning strategy on the checkpoints for another 40K steps. We plot the entity translation translation accuracy and BLEU of the test data in Figure~\ref{fig:pretrain_vs_perf}. We find that the checkpoint at 25K steps has already achieved a comparable entity translation accuracy with respect to the checkpoint at 150K steps. This shows that \method is efficient to learn the entity translations as early as in 25K steps. Besides, both the BLEU and entity translation accuracy keep improving as the pre-training steps increase to 200K steps.

\begin{table*}[th]
\centering \setlength{\tabcolsep}{2pt}
\resizebox{0.99\textwidth}{!}{%
    \begin{tabular}{rl} \toprule
        Src: & These new format stores have opened for business in \boldblue{Krasnodar}, \boldblue{Saratov}, and \boldblue{Ulyanovsk}. \\ 
        Ref: & \foreignlanguage{russian}{Магазины нового формата заработали в Краснодаре, Саратове и Ульяновске.} \\ \midrule
        \circled{1}  & \foreignlanguage{russian}{Эти новые форматовые магазины открылись для бизнеса в \boldred{Анридаре}, \boldred{Кристофе} и \boldred{Куьянме}.} \\
        \circled{2} & \foreignlanguage{russian}{Эти новые формат @-@ магазины открылись для бизнеса в \boldblue{Краснодаре}, \boldred{Сараабане} и в \boldred{Уругянском университете}.} \\
        \circled{3}  & \foreignlanguage{russian}{Эти новые магазины форматов открылись для бизнеса в \boldred{Krasnodar}, \boldred{Saratov} и \boldred{Ulyanovsk}.} \\
        \circled{4} & \foreignlanguage{russian}{Эти новые форматные магазины открылись для бизнеса в \boldblue{Краснодаре}, \boldblue{Саратове} и \boldblue{Ульяновске}.} \\ \midrule \midrule
        Src: & In \boldblue{Barnaul}, the new \boldblue{asphalt} on \boldblue{Krasnoarmeyskiy} Prospekt is being dug up \\  
        Ref: & \foreignlanguage{russian}{В \boldblue{Барнауле} вскрывают новый \boldblue{асфальт} на проспекте \boldblue{Красноармейском}} \\ \midrule
        \circled{1} & \foreignlanguage{russian}{В \boldblue{Барнауле} новое, как \boldred{разворачивающееся} на \boldred{железнополярном} Происсе, растет.}\\
        \circled{2} & \foreignlanguage{russian}{В \boldred{Барнале}, новое, как \boldred{разразилось} на \boldred{Красно @-@ Молгскиском} Просвещении, растет.}\\
        \circled{3} & \foreignlanguage{russian}{\boldred{Барнаул}, новый миф на \boldred{Krasnoarmey} Prospekt, выращивающий Krasnoarmeski.}\\
        \circled{4} &  \foreignlanguage{russian}{В \boldblue{Барнауле} новый \boldblue{асфальт} на \boldblue{Красноармейском} проспекте выращивание растет.}\\
        \bottomrule
    \end{tabular}
    }
    \caption{Qualitative comparison among four pre-training methods on named entity translations. \circled{1}: DAE $\rightarrow$ MT, \circled{2}: DEEP $\rightarrow$ MT, \circled{3}: DAE $\rightarrow$ DAE+MT, \circled{4}: DEEP $\rightarrow$ DEEP+MT. }
    \label{tab:my_label}
    \vspace{-1mm}
\end{table*}

\subsection{Qualitative Analysis}
In this section, we select two examples that contain entities appearing only in the pre-training and testing data. The first example contains three location names. We find that the model trained by the single-task DAE predicts the wrong places which provide the wrong information in the translated sentence. In addition, the model trained by the multitask DAE just copies the English named entities (i.e., ``\blue{Krasnodar}'', ``\blue{Saratov}'' and ``\blue{Ulyanovsk}'') to the target sentence without actual translation. In contrast, our method predicts the correct translation for ``\blue{Krasnodar}'' in both single-task and multi-task setting, while the multi-task \method translates all entities correctly. In the second example, although our method in the single-task setting predicts wrong for all the entities, the model generates partially correct translations such as ``\foreignlanguage{russian}{\red{Барнале}}'' for ``\foreignlanguage{russian}{\blue{Барнауле}}'' and ``\foreignlanguage{russian}{\red{Красно @-@ Молгскиском}}'' for ``\foreignlanguage{russian}{\blue{Красноармейском}}''. Notice that \method in the multi-task setting translates the correct entities ``\blue{asphalt}'' and ``\blue{Krasnoarmeyskiy}'' which convey the key information in this sentence. In contrast, the translation produced by the multi-task DAE method literally means ``\foreignlanguage{russian}{\red{Барнаул} (Barnaul), новый (new) миф (myth) на (at) \red{Krasnoarmey} Prospekt, выращивающий (grow) Krasnoarmeski.}'', which is incomprehensible due to the entity translation errors.

\section{Related Work}

\textbf{Named Entity Translation} has been extensively studied for decades~\cite{arbabi94,knight-graehl-1998-machine}. Earlier studies focus on rule-based methods using phoneme or grapheme~\cite{wan-verspoor-1998-automatic,al-onaizan-knight-2002-translating}, statistical methods that align entities in parallel corpus~\cite{huang-etal-2003-automatic,huang-etal-2004-improving,zhang-etal-2005-phrase} and Web mining methods built on top of a search engine~\cite{huang-etal-2005-mining,wu-chang-2007-learning,yang-etal-2009-chinese}. Recently, neural models have been applied for named entity translations. \citet{finch-etal-2016-target,HADJAMEUR2017287,grundkiewicz-heafield-2018-neural} used neural machine translation to transliterate named entities. \citet{ugawa-etal-2018-neural,DBLP:journals/corr/abs-2009-13398} integrated named entity tags to neural machine translation models. In this paper, without changing model architectures, we focus on data augmentation methods to improve name entity translation within context.

\noindent\textbf{Pre-training of Neural Machine Translation} has been shown effective in low-resource and medium-resource language translations by many recent works~\cite{NEURIPS2019_c04c19c2,pmlr-v97-song19d,liu-etal-2020-multilingual-denoising,lin-etal-2020-pre}, where different pre-training objectives are proposed to leverage large amounts of monolingual data for translation. These methods adopt a denoising auto-encoding framework, which encompasses several different works in data augmentation on monolingual data for MT~\cite{lambert-etal-2011-investigations,currey-etal-2017-copied,sennrich-etal-2016-improving,hu-etal-2019-domain-adaptation}. However, named entity translations during the pre-training is under-explored. We fill this gap by integrating named entity recognition and linking to the pre-training of neural machine translation. Moreover, while recent work shows that continue finetuning a pre-trained encoder with the same pre-training objective improves language understanding tasks~\cite{gururangan-etal-2020-dont}, this finetuning paradigm has not been explored for pre-training of a sequence-to-sequence model. Besides, previous works on multitask learning for MT focus on language modeling~\cite{gulcehre2015using,zhang-zong-2016-exploiting,domhan-hieber-2017-using,zhou-etal-2019-improving}, while we examine a multi-task finetuning strategy with an entity-based denoising task in this work and demonstrate substantial improvements for named entity translations.

\section{Conclusion}
In this paper, we propose an entity-based pre-training method for neural machine translation. Our method improves named entity translation accuracy as well as BLEU score over strong denoising auto-encoding baselines in both single-task and multi-task setting. Despite the effectiveness, several challenging and promising directions can be considered in the future. First, recent works on integrating knowledge graphs~\cite{zhao-etal-2020-knowledge-graph,ijcai2020-559} in neural machine translation have shown promising results for translation. Our method links entities to a multilingual knowledge base which contains rich information of the entities such as entity description, relation links, alias. How to leverage these richer data sources to resolve entity ambiguity deserves further investigation. Second, finetuning pre-trained models on in-domain text data is a potential way to improve entity translations across domains.

\bibliography{reference_short}
\bibliographystyle{acl_natbib}

\clearpage
\newpage
\onecolumn
\appendix
\section*{Appendix}

\section{Finetuning BLEU Curves} \label{sec:stepwise_finetune}
\begin{figure*}[h]
    \centering
    \includegraphics[width=\textwidth]{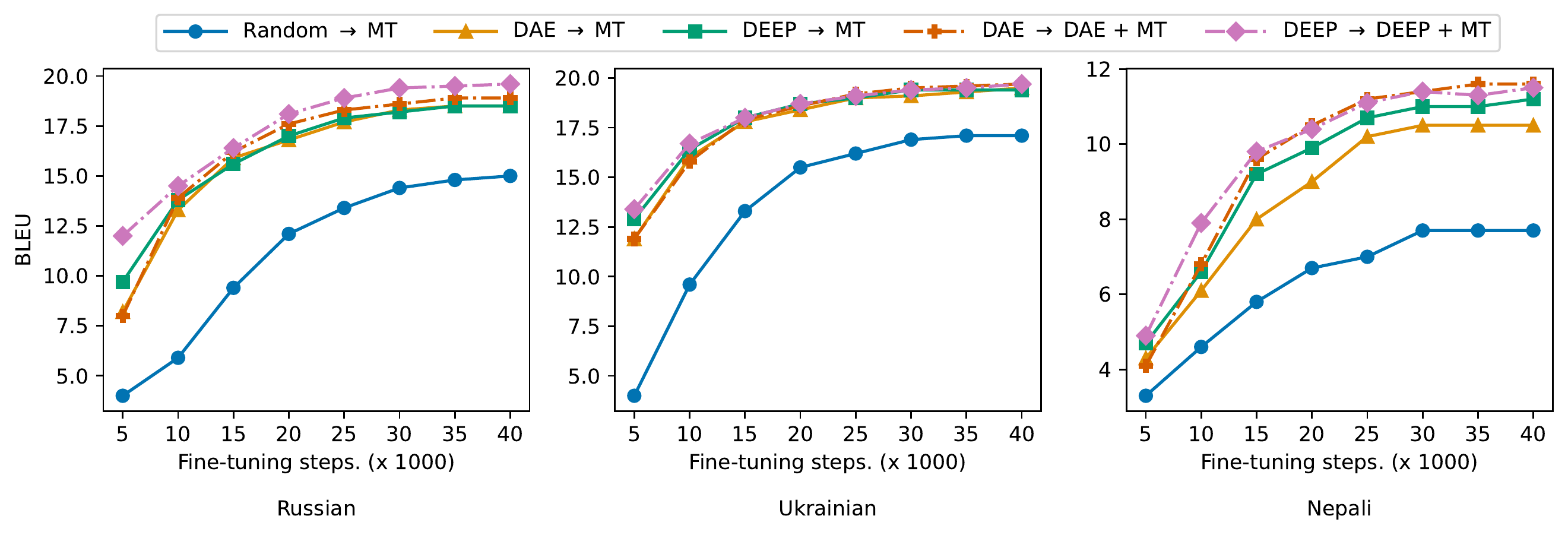}
    \caption{\textbf{BLEU} scores for 3 language pairs over various finetuning steps.}
    \label{fig:per_step_bleu}
\end{figure*}

We report BLEU score for three language pairs calculated from  checkpoints at different finetuning steps in Figure~\ref{fig:per_step_bleu}.
For all language pairs, all pre-training methods result in a significant increase in terms of BLEU throughout the finetuning in both single-task and multi-task setting.
In particular, the differences in BLEU between \method and the other baselines are most significant at the beginning of the finetuning stage.

\section{Entity Translation Accuracy for other languages}
\label{app:eta_for_uk_ne}
\begin{figure*}
    \centering
    \includegraphics[width=\textwidth]{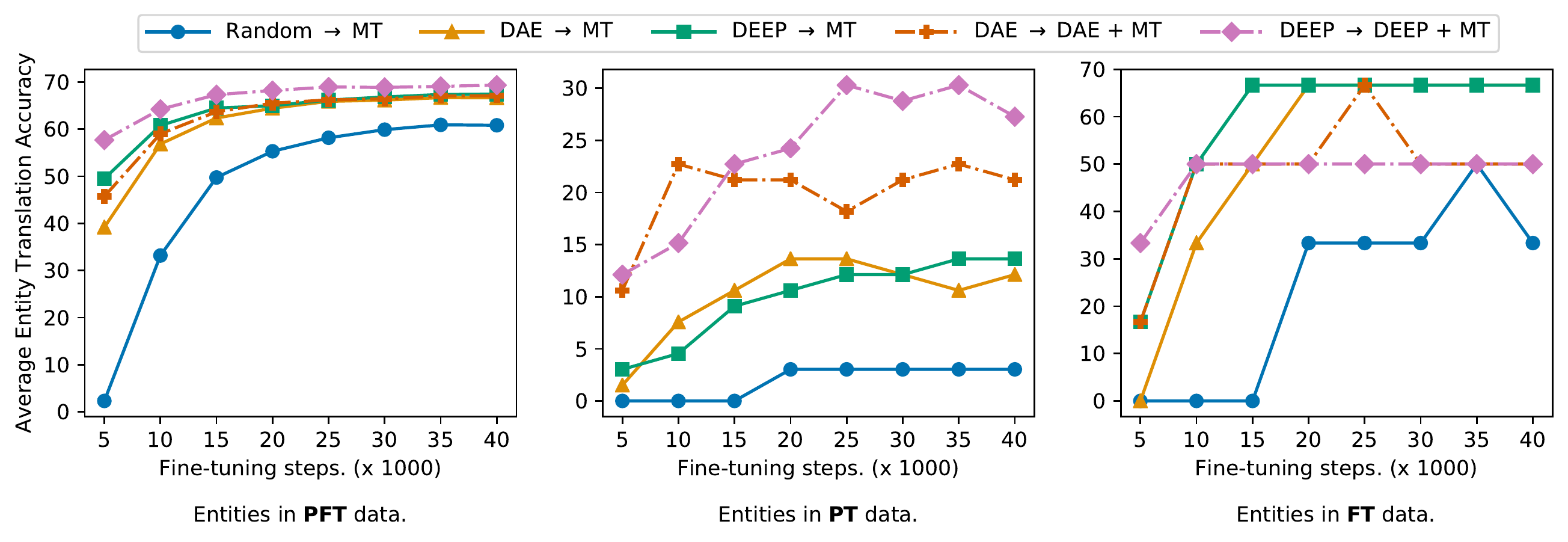}
    \caption{\textbf{Entity translation accuracy} aggregated over different entity sets for \textbf{Ukrainian}.}%
    \label{fig:uk_stepwise}
\end{figure*}

\begin{figure*}
    \centering
    \includegraphics[width=\textwidth]{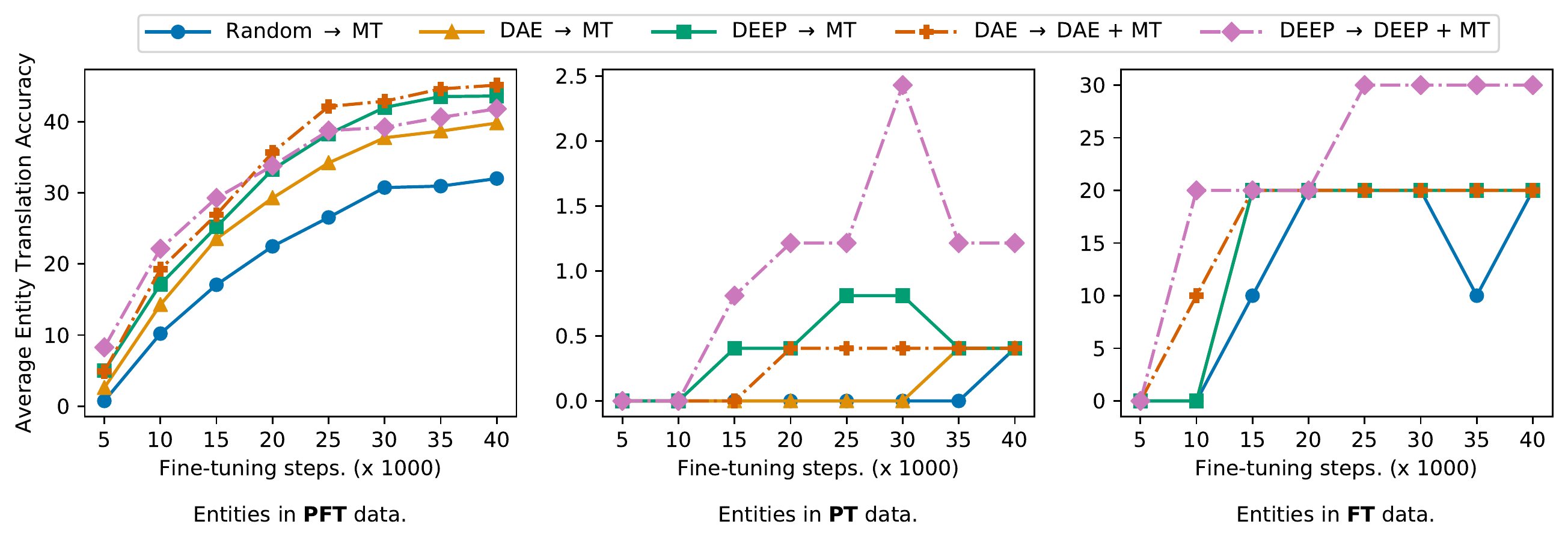}
    \caption{\textbf{Entity translation accuracy} aggregated over different entity sets for \textbf{Nepali}.}%
    \label{fig:ne_stepwise}
\end{figure*}

\begin{figure*}
    \centering
    \includegraphics[width=\textwidth]{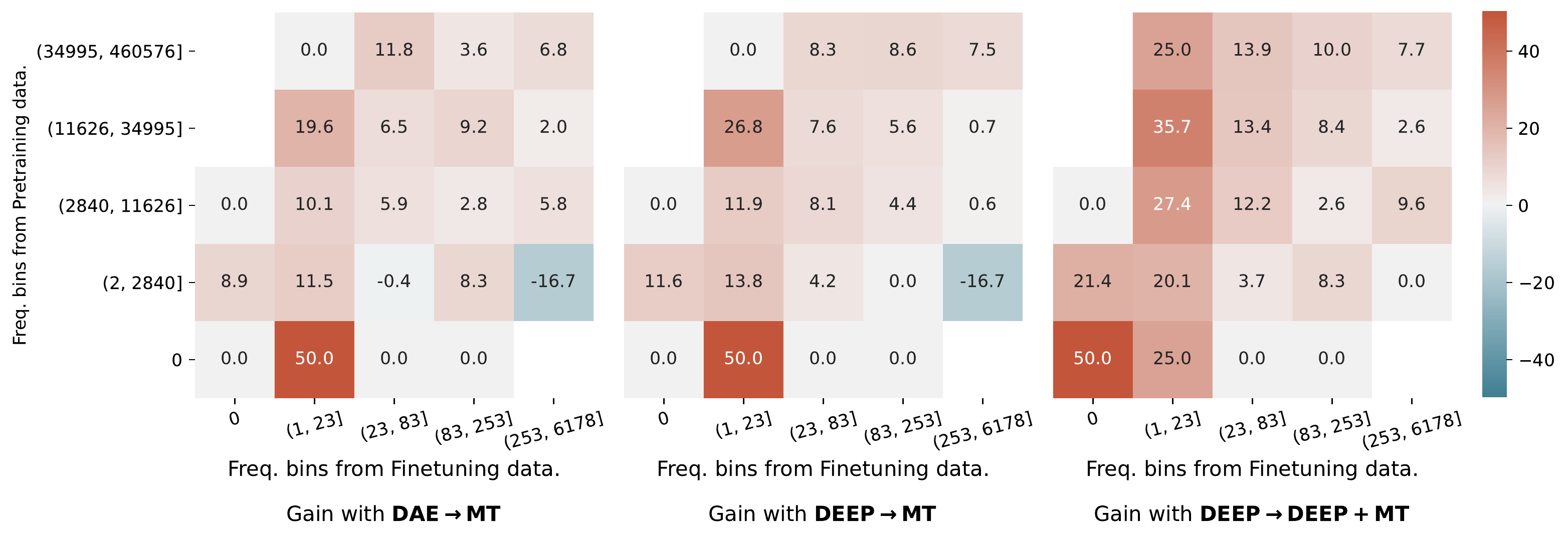}
    \caption{Gain from \textbf{Random $\rightarrow$ MT} in entity translation accuracy for \textbf{Ukrainian} for each model.}%
    \label{fig:uk_heatmap}
\end{figure*}

\begin{figure*}
    \centering
    \includegraphics[width=\textwidth]{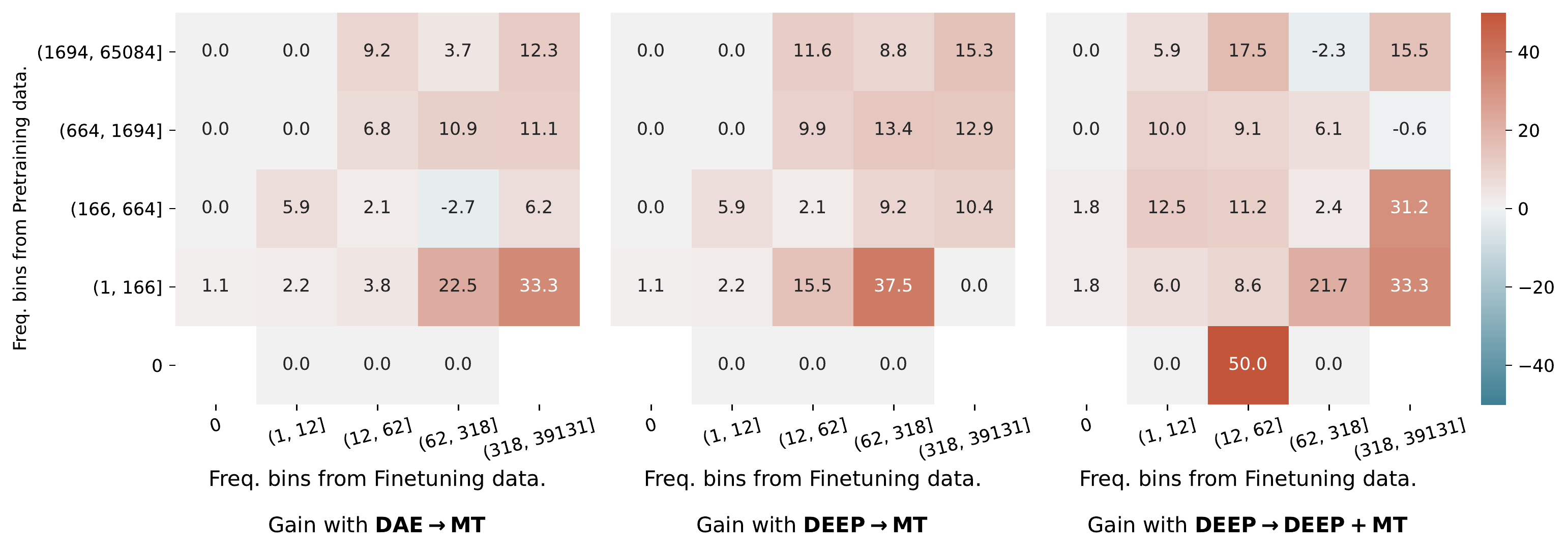}
    \caption{Gain from \textbf{Random $\rightarrow$ MT} in entity translation accuracy for \textbf{Nepali} for each model.}%
    \label{fig:ne_heatmap}
\end{figure*}

We show the entity translation accuracy performance over various finetuning steps for Ukrainian and Nepali in Figure~\ref{fig:uk_stepwise}, \ref{fig:ne_stepwise}, and show the gains of three pre-training methods over the random baseline with respect to the entity frequencies in Figure~\ref{fig:uk_heatmap}, \ref{fig:ne_heatmap}.
Empty cells in the heatmaps are due to no entities that meet the conditions in those cells.

\paragraph{Ukrainian:}
As seen in Figure~\ref{fig:uk_stepwise}, the general trend for the entity translation accuracy according to entity groups are similar to that of Russian.
While \method achieves the highest accuracy in \textbf{FT}, the results for \textbf{FT} is less reliable due to a small sample size of entities in \textbf{FT}.
In terms of the gain from \textbf{Random $\rightarrow$ MT} according to the entity frequency, we observe a consistent improvement of our multi-task \method on translating low-frequent entities in the finetuning data (See the left bottom of Figure~\ref{fig:uk_heatmap}).

\paragraph{Nepali:}
While outperforming at the beginning of finetuning, Figure~\ref{fig:ne_stepwise} shows that \textbf{DEEP $\rightarrow$ DEEP+MT} eventually under-performed for translations of entities in \textbf{PFT} data.
Moreover, the accuracy is considerably lower on entities in \textbf{PT}, which suggests that the degree of forgetting is much more conspicuous in Nepali.
The gain from \textbf{Random $\rightarrow$ MT} with respect to the entity frequency exhibited a different trend from Russian and Ukrainian.
Figure~\ref{fig:ne_heatmap} shows the results.
In the single-task setting, \method improve the translations of frequent entities appearing in both the pre-training and finetuning data.
Despite the multi-task learning that introduces additional exposure to entities that are more frequent in the pre-training data, the largest gain comes from entities that are less frequent in the pre-training data but frequent in the finetuning data.

\end{document}